\documentclass[twocolumn]{article}

\usepackage{mystyle}
\usepackage{times}  
\usepackage{helvet} 
\usepackage{courier}  
\usepackage[hyphens]{url}  
\usepackage{graphicx} 
\usepackage{bm}
\usepackage{xcolor}
\usepackage{graphicx}  
\def\argmax{\mathop{\rm argmax}}%
\def\argmin{\mathop{\rm argmin}}%

\usepackage{multirow}
\usepackage{algpseudocode}

\title{Neural Machine Translation with Explicit Phrase Alignment}
\date{}

\author{\textbf{Jiacheng Zhang$^\dagger$,  Huanbo Luan$^\dagger$, Maosong Sun$^{\dagger}$, FeiFei Zhai$^\#$,} \\ \textbf{Jingfang Xu$^\#$  and Yang Liu$^{\dagger\ddagger}$} \\
$^\dagger$Institute for Artificial Intelligence \\
    State Key Laboratory of Intelligent Technology and Systems \\
    Department of Computer Science and Technology, Tsinghua University, Beijing, China \\
    $^\ddagger$Beijing National Research Center for Information Science and Technology\\
    $^\#$Sogou Inc., Beijing, China\\
}

\begin{document}

\maketitle

\begin{abstract}
While neural machine translation (NMT) has achieved state-of-the-art translation performance, it is unable to capture the alignment between the input and output during the translation process. The lack of alignment in NMT models leads to three problems: it is hard to (1) interpret the translation process, (2) impose lexical constraints, and (3) impose structural constraints. To alleviate these problems, we propose to introduce explicit phrase alignment into the translation process of arbitrary NMT models. The key idea is to build a search space similar to that of phrase-based statistical machine translation for NMT where phrase alignment is readily available. We design a new decoding algorithm that can easily impose lexical and structural constraints. Experiments show that our approach makes the translation process of NMT more interpretable without sacrificing translation quality. In addition, our approach achieves significant improvements in lexically and structurally constrained translation tasks.
\end{abstract}

\section{Introduction}
Neural machine translation (NMT), which leverages neural networks to map between natural languages, has made remarkable progress in the past several years \cite{Sutskever:14,Bahdanau:15,Vaswani:17}. Capable of learning representations from data, NMT has achieved significant improvements over conventional statistical machine translation (SMT) \cite{Koehn:03} and become the new {\em de facto} paradigm in the machine translation community.

Despite its success, NMT suffers from a major drawback: there is no alignment to explicitly indicate the correspondence between the input and the output. As all internal information of an NMT model is represented as real-valued vectors or matrices, it is hard to associate a source word with its translational equivalents on the target side. Although the attention weights between the input and the output are available in the RNNsearch model \cite{Bahdanau:15}, these weights only reflect relevance rather than translational equivalence \cite{Koehn:17}. To aggravate the situation, attention weights between the input and the output are even unavailable in modern NMT models such as Transformer \cite{Vaswani:17}.

The lack of alignment in NMT leads to at least three problems. First, it is difficult to interpret the translation process of NMT models without alignment. In conventional SMT \cite{Koehn:03}, the translation process can be seen as a sequence of interpretable decisions, in which alignment plays a central role. It is hard to include such interpretable decisions in NMT models without the access to alignment. Although visualization tools such as layer-wise relevance propagation \cite{Ding:17} can be used to measure the relevance between two arbitrary neurons in NMT models, the hidden states in neural networks still do not have clear connections to interpretable language structures.

Second, it is difficult to impose \textbf{lexical constraints} on NMT systems \cite{Hokamp:17} without alignment. For example, given an English sentence
\begin{center}
\verb|American peot Edgar Allan Poe|
\end{center}
, one requires that the English phrase ``Edgar Allan Poe'' must be translated by NMT systems as a Chinese word ``ailunpo''. Such lexical constraints are important for both automatic MT and interactive MT. In automatic MT, it is desirable to incorporate the translations of infrequent numbers, named entities, and technical terms into NMT systems \cite{Luong:15}. In interactive MT, human experts expect that the NMT system can be controlled and include specified translations in the system output \cite{Cheng:16}. Although \citet{Hokamp:17} and \citet{Post:18} provide solutions to impose lexical constraints, their methods can only ensure that the specified target words or phrases will appear in the system output. As a result, the ignorance of the alignment to the source side might deteriorate the adequacy of system output (see Table \ref{tab:lex_con}).

\begin{figure}[!t]
\centering
\includegraphics[width=0.4\textwidth]{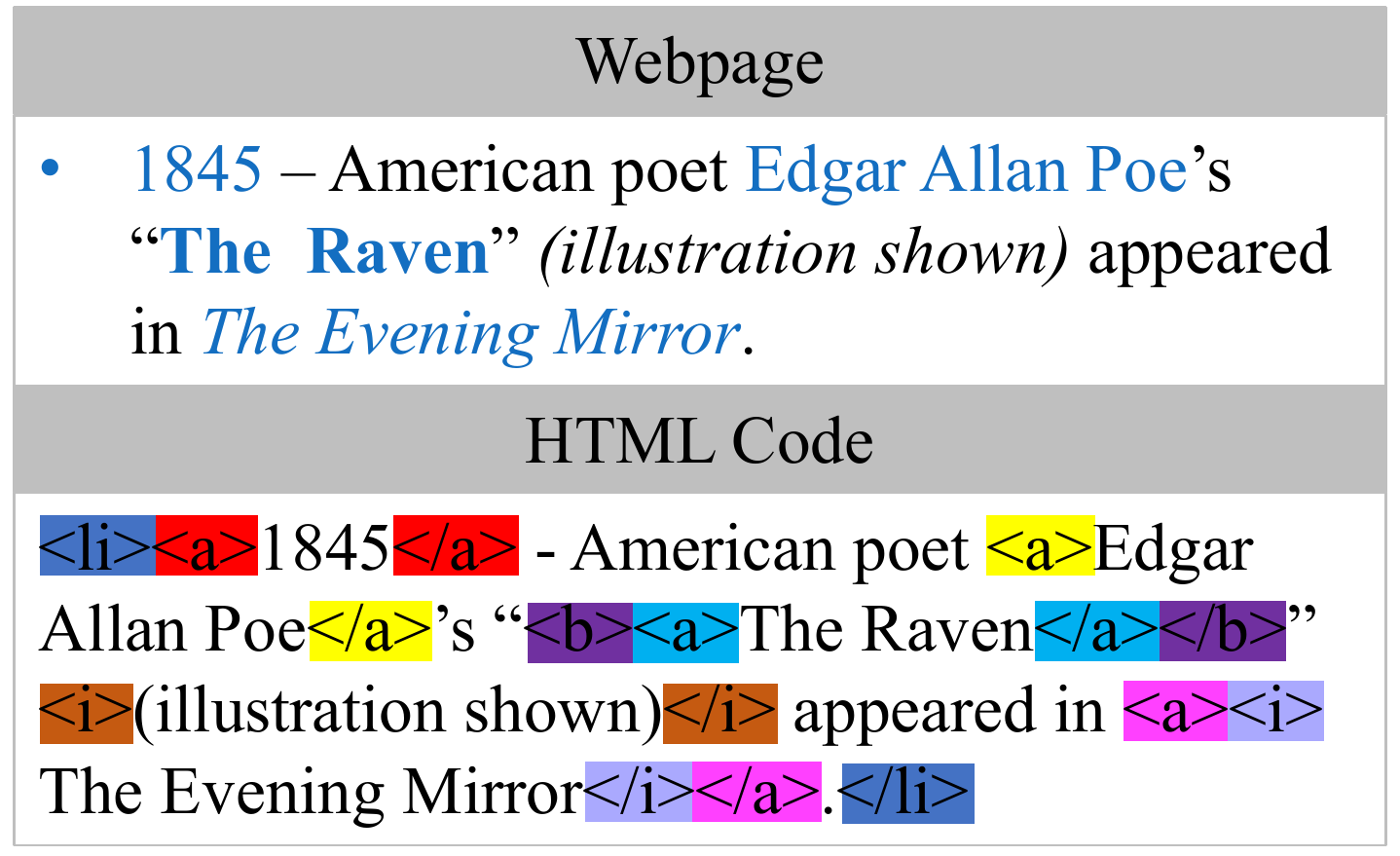}
\caption{Example structural constraints. The translation of a source string enclosed in a pair of HTML tags must be confined by the same tag pair on the target side.} \label{fig:example_structured_text}
\end{figure}

Third, it is difficult to impose \textbf{structural constraints} on NMT systems without alignment. Figure \ref{fig:example_structured_text} shows an example of webpage and its HTML code. Unlike lexical constraints, structural constraints require that source strings enclosed in paired HTML tags must be translated as single units and the translations must be enclosed in the same paired HTML tags. For example, the Chinese translation of ``$\langle$a$\rangle$The Raven$\langle$/a$\rangle$'' should be ``$\langle$a$\rangle$wuya$\langle$/a$\rangle$''. It is challenging for NMT models trained on plain text to translate such structured text. While removing these HTML tags before translation and inserting tags back after translation will maintain translation quality but often violate structural constraints \cite{Zantout:01,Joanis:13}, only translating the plain text within tags and concatenating the translations and tags in a monotonic way can strictly conform to structural constraints but impair translation quality \cite{Al-Anzi:97}.

In this work, we propose to introduce phrase alignment into the translation process of arbitrary NMT models. The basic idea is to develop an NMT model that treats phrase alignment as a latent variable. During decoding, the NMT model is used to score a search space similar with conventional phrase-based SMT \cite{Koehn:03}, in which phrase alignment is readily available. While the use of the trained NMT model keeps the capabilities of NMT in learning representations from data and capturing non-local dependencies, the availability of phrase alignment makes it possible to include interpretable decisions in the translation process. Also thanks to the availability of phrase alignment, we design a new decoding algorithm that applies to all the unconstrained, lexically constrained, and structurally constrained translation tasks. Experiments show that the use of phrase-based search space does not hurt the translation performance of NMT models on the unconstrained translation task. Moreover, our approach significantly improves over state-of-the-art methods on the lexically and structurally constrained translation tasks.

\section{Related Work}

Our work is related to three lines of research: (1) interpreting NMT, (2) constrained decoding for NMT, and (3) combining SMT and NMT.

\subsection{Interpreting NMT}

Our work is related to attempts on interpreting NMT \cite{Ding:17,li:19}. \citet{Ding:17} calculate the relevance between source and target words with layer-wise relevance propagation. Such relevance measures the contribution of each source word to target word instead of translational equivalence between source and target words. \citet{li:19} predict alignment with an external alignment model trained on the output of a statistical word aligner and use prediction differences to quantify the relevance between source and target words. However, their external alignment model is not identical to the alignment in the translation process. Our approach differs from prior studies by introducing explicit phrase alignment into the translation process of NMT models, which makes each step in generating a target sentence interpretable to human experts.

\subsection{Constrained Decoding for NMT} \label{sec:wt}

Our work is also closely related to imposing lexical constraints on the decoding process of NMT \cite{Hokamp:17,Chatterjee:17,Post:18}. \citet{Hokamp:17} propose a lexically constrained decoding algorithm for NMT. Their approach can ensure that pre-specified target strings will appear in the system output. \citet{Post:18} improve the efficiency of lexically constrained decoding by introducing dynamic beam allocation. One drawback of the two methods is that they cannot impose lexical constraints on the source side due to the lack of alignment. \citet{Chatterjee:17} and \citet{hasler:18} rely on the attention weights in the RNNsearch model \cite{Bahdanau:15} to impose source-aware lexical constraints with guided beam search. However, their methods can not be applied to Transformer \cite{Vaswani:17}. With translation options, it is also easy to impose source-aware lexical constraints using our approach for arbitrary NMT models. 

The direction of imposing structural constraints remains much unexplored, especially for NMT. Most prior studies have focused on SMT. Although the ideal solution is to directly train NMT models on parallel corpora for structured text \cite{Du:10,Hudic:11,Tezcan:11}, such labeled datasets are hard to construct and remain limited in quantity. Therefore, a more practical solution is to use off-the-shelf MT systems tailored for unstructured text to translate structured text \cite{Al-Anzi:97,Zantout:01,Joanis:13}. But these approaches face the risk of performance degradation or failure to impose structural constraints correctly. Our work proposes a structurally constrained decoding algorithm for NMT to preserve structural constraints without sacrificing translation quality. 

\subsection{Combining SMT and NMT}

Several authors have endeavored to combine the merits of SMT and NMT \cite{Stahlberg:16,Khayrallah:17,Dahlmann:17}. \citet{Stahlberg:16} propose to use the lattice output by SMT as the search space of NMT. The major difference is that our work allows for both source word omission and target word insertion, which seem to be helpful in reducing the gap between phrase-based and neural spaces. In this work, we only use NMT models to score the translations in a phrase-based space. It is possible to exploit SMT features as suggested by \citet{Dahlmann:17}.


\section{Approach}

\begin{figure}[!t]
\centering
\includegraphics[width=0.35\textwidth]{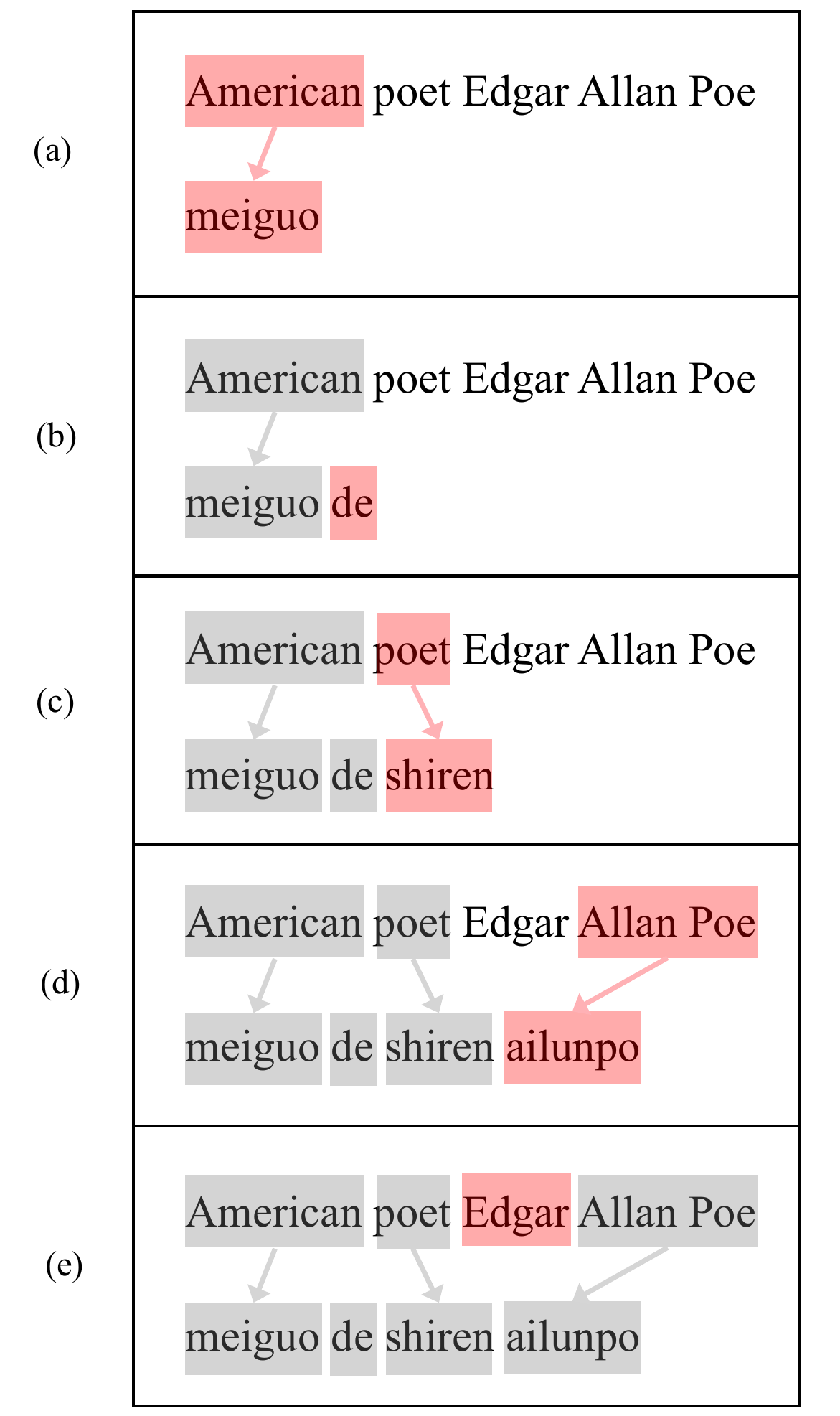}
\caption{Neural machine translation with explicit phrase alignment.} \label{fig:phrase_alignment}
\end{figure}

Our work aims to introduce phrase alignment into the translation process of arbitrary NMT models. Figure \ref{fig:phrase_alignment} illustrate the central idea of our approach. During decoding, the target sentence and phrase alignment are generated simultaneously. As the target sentence grows from left to right, it is easy to apply arbitrary NMT models to calculate translation probabilities in an incremental way. A key difference of our approach from conventional phrase-based SMT \cite{Koehn:03} is that unaligned source and target phrases are allowed to reduce the discrepancy between the search spaces of SMT and NMT models. For example, Figure \ref{fig:phrase_alignment}(b) uses an unaligned target phrase (i.e., ``de'') and Figure \ref{fig:phrase_alignment}(e) uses an unaligned source phrase (e.g., ``Edgar''). With access to phrase alignment, we develop a decoding algorithm that is capable of preserving lexical and structural constraints without sacrificing translation quality. 

\subsection{Modeling}

Let $\mathbf{x}=x_1, \dots, x_I$ be a source sentence and $\mathbf{y}=y_1, \dots, y_J$ be a target sentence. We use $x_0$ to denote an empty source word that connects to all unaligned target phrases and $y_0$ to denote an empty target word that connects to all unaligned source phrases. We use $\mathbf{z} = z_1, \dots, z_K$ to denote the phrase alignment between the source and target sentences. Each link $z_k = (i_b, i_e, j_b, j_e)$ is a 4-tuple, where $i_b$ is the beginning position of the source phrase, $i_e$ is the ending position of the source phrase, $j_b$ is the beginning position of the target phrase, and $j_e$ is the ending position of the target phrase. For example, the phrase alignment in Figure \ref{fig:phrase_alignment} comprises five links: $z_1 = (1, 1, 1, 1)$, $z_2 = (0, 0, 2, 2)$, $z_3 = (2, 2, 3, 3)$, $z_4 = (4, 5, 4, 4)$, and $z_5 = (3, 3, 0, 0)$. For convenience, we use $\mathbf{x}_{z_k}$ to denote the source phrase spanning from $i_b$ to $i_e$ and $\mathbf{y}_{z_k}$ to denote the target phrase spanning from $j_b$ to $j_e$. For example, $\mathbf{x}_{z_4}$ is ``Allan Poe'' and $\mathbf{y}_{z_4}$ is ``alunpo''.

More formally, our approach is based on a latent variable model given by
\begin{eqnarray}
P(\mathbf{y}|\mathbf{x}; \bm{\theta}) &=& \sum_{\mathbf{z}} P(\mathbf{y}, \mathbf{z} | \mathbf{x}; \bm{\theta}),
\end{eqnarray}
where $\bm{\theta}$ is a set of model parameters.

The probability of generating the target sentence $\mathbf{y}$ and phrase alignment $\mathbf{z}$ given the source sentence $\mathbf{x}$ can be further factored as
\begin{eqnarray}
P(\mathbf{y}, \mathbf{z}|\mathbf{x}; \bm{\theta}) = \prod_{k=1}^{K} P(z_k | \mathbf{x}, \mathbf{y}_{z_1}, \dots, \mathbf{y}_{z_{k-1}}, \mathbf{z}_{<k}; \bm{\theta}) \nonumber \\
 P(\mathbf{y}_{z_k} | \mathbf{x}, \mathbf{y}_{z_1}, \dots, \mathbf{y}_{z_{k-1}}, \mathbf{z}_k; \bm{\theta}), \quad \
\end{eqnarray}
where $P(z_k | \mathbf{x}, \mathbf{y}_{z_1}, \dots, \mathbf{y}_{z_{k-1}}, \mathbf{z}_{<k}; \bm{\theta})$ is a {\em phrase alignment} model and $P(\mathbf{y}_{z_k} | \mathbf{x}, \mathbf{y}_{z_1}, \dots, \mathbf{y}_{z_{k-1}}; \bm{\theta})$ is a {\em phrase translation} model. Note that $\mathbf{z}_{<k}=z_1, \dots z_{k-1}$ is a partial phrase alignment. As it is challenging to estimate the phrase alignment model from data due to the exponential search space of phrase alignments, we assume that the alignment model has a uniform distribution for simplicity and leave the learning of the alignment model for future work.

We distinguish between two kinds of phrase translation models: {\em non-empty} and {\em empty}. For non-empty target phrases, the phrase translation probability can be decomposed as a product of word-level translation probabilities:
\begin{eqnarray}
P(\mathbf{y}_{z_k} | \mathbf{x}, \mathbf{y}_{z_1}, \dots, \mathbf{y}_{z_{k-1}}; \bm{\theta}) \quad \quad \quad \quad \quad \quad \quad \quad \quad \ \ \nonumber \\
= \prod_{l=1}^{|\mathbf{y}_{z_k}|} P(\mathbf{y}_{z_k}^{(l)} | \mathbf{x}, \mathbf{y}_{z_1}, \dots, \mathbf{y}_{z_{k-1}}, \mathbf{y}_{z_k}^{(1)}, \dots, \mathbf{y}_{z_k}^{(l-1)}; \bm{\theta}_n), \label{eq:non-empty}
\end{eqnarray}
where $\mathbf{y}_{z_k}^{(l)}$ is the $l$-th word in the target phrase $\mathbf{y}_{z_k}$ and $\bm{\theta}_n$ denotes the set of model parameters related to non-empty phrases. Note that the word-level translation probabilities in Eq. (\ref{eq:non-empty}) can be easily calculated by arbitrary NMT models.

For the empty target phrase such as $\mathbf{y}_{z_5}= y_0$, we define the phrase translation probability as
\begin{eqnarray}
&& P(\mathbf{y}_{z_k} | \mathbf{x}, \mathbf{y}_{z_1}, \dots, \mathbf{y}_{z_{k-1}}; \bm{\theta}) \nonumber \\
&=& P(y_0 | \mathbf{x}_{z_k}, \mathbf{x} / \mathbf{x}_{z_k}; \bm{\theta}_e), \label{eq:empty}
\end{eqnarray}
where $\mathbf{x}_{z_k}$ is the source phrase aligned to $y_0$, $\mathbf{x} / \mathbf{x}_{z_k}$ is the surrounding context on the source side, and $\bm{\theta}_{e}$ is the set of model parameters related to empty phrases. For simplicity, we restrict that unaligned source phrase $\mathbf{x}_{z_k}$ to be a single source word. Note that $\bm{\theta} = \bm{\theta}_n \cup \bm{\theta}_e$.

We use the self-attention based encoder \cite{Vaswani:17} to model the translation probability of empty target phrases. The encoder takes $\mathbf{x}_{z_k}$ and $\mathbf{x} / \mathbf{x}_{z_k}$ as input and output the probability of omitting $\mathbf{x}_{z_k}$.

\begin{figure*}[!t]
\centering
\includegraphics[width=0.75\textwidth]{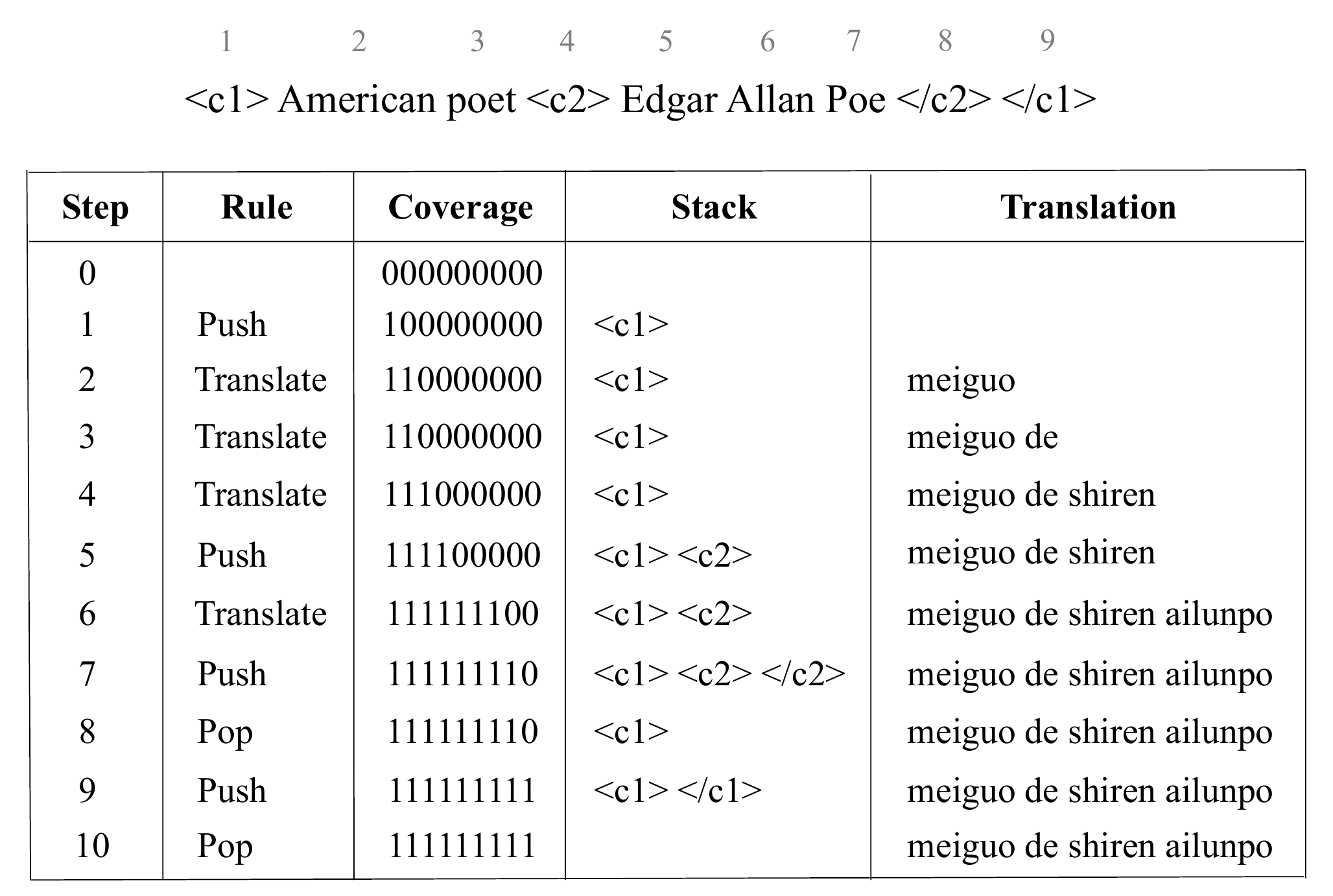}
\caption{An example derivation of structurally constrained decoding.} \label{fig:derivation}
\end{figure*}

\begin{figure}[!t]
\centering
\includegraphics[width=0.45\textwidth]{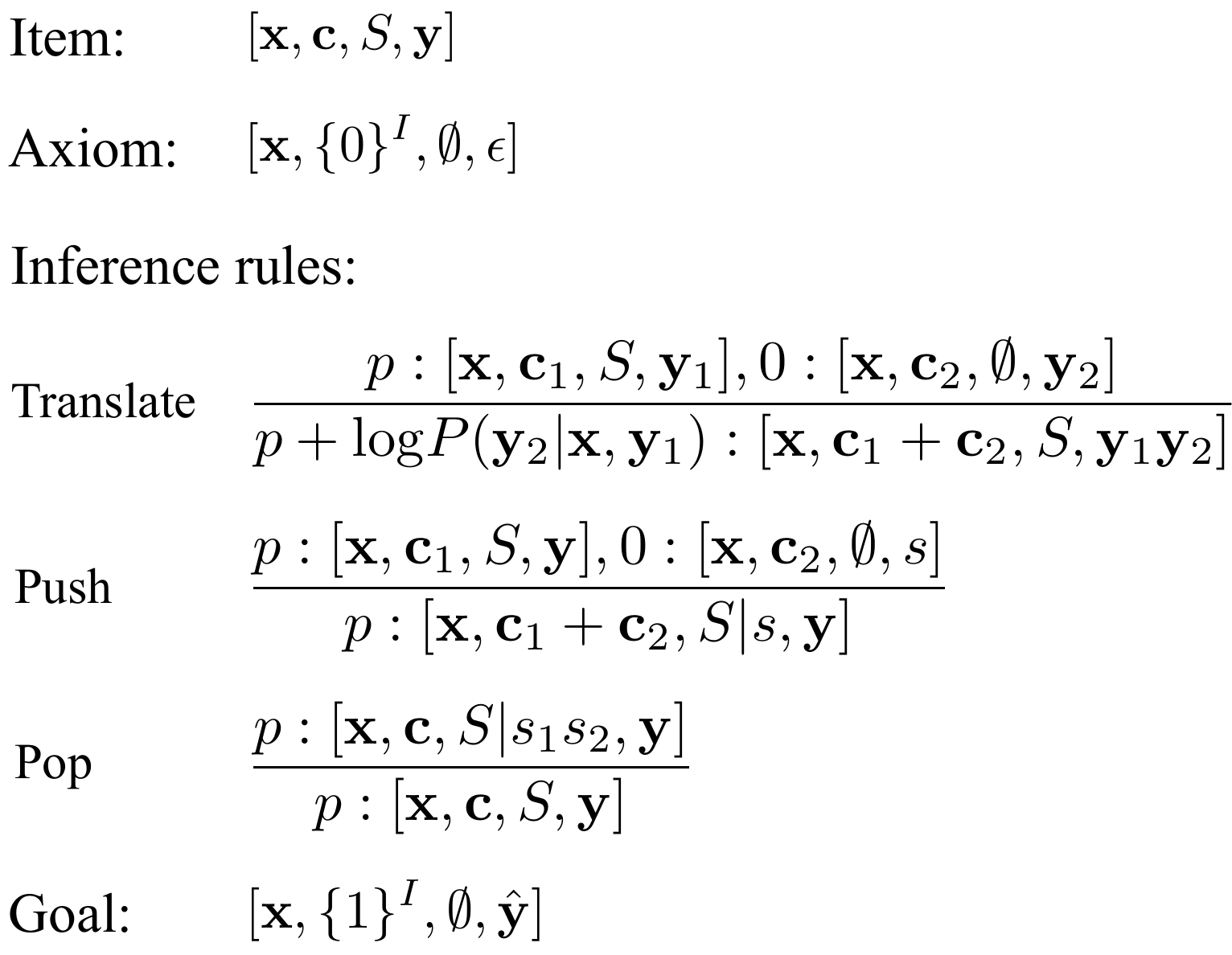}
\caption{The deductive system of structurally constrained decoding.} \label{fig:deductive}
\end{figure}

\subsection{Training}

Given a parallel corpus $D = \{ \langle \mathbf{x}^{(s)}, \mathbf{y}^{(s)} \rangle \}_{s=1}^{S}$, the standard training objective is to maximize the log-likelihood of the training data:
\begin{eqnarray}
\hat{\bm{\theta}} = \argmax_{\bm{\theta}} \Bigg\{ \sum_{s=1}^{S} \log P(\mathbf{y}^{(s)} | \mathbf{x}^{(s)}; \bm{\theta}) \Bigg\}.
\end{eqnarray}

As training the latent-variable model requires to enumerate all possible phrase alignments, it is impractical to directly estimate $\bm{\theta}_n$ and $\bm{\theta}_e$ jointly. Instead, we propose to train the two models separately. For the non-empty translation model in Eq. (\ref{eq:non-empty}), the training objective is given by
\begin{eqnarray}
\hat{\bm{\theta}}_n \quad \quad \quad \quad \quad \quad \quad \quad \quad \quad \quad \quad \quad \quad \quad \quad \quad \quad \ \nonumber \\ 
= \argmax_{\bm{\theta}_n} \Bigg\{ \sum_{s=1}^{S} \sum_{j=1}^{|\mathbf{y}^{(s)}|} \log P(y^{(s)}_j | \mathbf{x}^{(s)}, \mathbf{y}^{(s)}_{<j}; \bm{\theta}_n) \Bigg\}.
\end{eqnarray}

For the empty translation model in Eq. (\ref{eq:empty}), we can use an external word alignment tool \cite{Och:03} to generate word alignments for the parallel corpus $D$. It is easy to decide whether a source word is unaligned or not based on the word alignments. As a result, the training objective for the empty translation model is given by
\begin{eqnarray}
\hat{\bm{\theta}}_e = \argmin_{\bm{\theta}_e} \Bigg\{ \sum_{s=1}^{S} \sum_{i=1}^{|\mathbf{x}^{(s)}|} \mathrm{CE}(\mathbf{x}^{(s)}, \mathbf{u}^{(s)}, \bm{\theta}_e, i) \Bigg\},
\end{eqnarray}
where $\mathbf{u}^{(s)} = u^{(s)}_1, \dots, u^{(s)}_I$ is an indicator vector corresponding to the $s$-th source sentence $\mathbf{x}^{(s)}$ that indicates whether $x^{(s)}_i$ is unaligned and $\mathrm{CE}(\cdot)$ is the cross entropy loss defined as
\begin{eqnarray}
&&\mathrm{CE}(\mathbf{x}^{(s)}, \mathbf{u}^{(s)}, \bm{\theta}_e, i) \nonumber \\
&=& -u^{(s)}_i \log P(y_0 | x^{(s)}_i, \mathbf{x}^{(s)} / x^{(s)}_i; \bm{\theta}_e) + \nonumber \\
&& (1 - u^{(s)}_i) \log \Big(1 - P(y_0 | x^{(s)}_i, \mathbf{x}^{(s)} / x^{(s)}_i; \bm{\theta}_e) \Big). \quad \
\end{eqnarray}

\subsection{Decoding}

Given the learned model parameters $\hat{\bm{\theta}} = \hat{\bm{\theta}}_n \cup \hat{\bm{\theta}}_e$ and an unseen source sentence $\mathbf{x}$, our goal is to find the target sentence $\hat{\mathbf{y}}$ and phrase alignment $\hat{\mathbf{z}}$ with the highest probability without violating pre-specified constraints:
\begin{eqnarray}
\hat{\mathbf{y}}, \hat{\mathbf{z}} = \argmax_{\mathbf{y}, \mathbf{z} \ s.t. \ C(\mathbf{x}, \mathbf{y}, \mathbf{z}, \mathcal{C})=1} \Big\{ P(\mathbf{y}, \mathbf{z} | \mathbf{x}; \hat{\bm{\theta}}) \Big\},
\end{eqnarray}
where $C(\mathbf{x}, \mathbf{y}, \mathbf{z}, \mathcal{C})$ is a function that checks whether the resulting translation and alignment conform to a set of pre-specified constraints $\mathcal{C}$. The function returns 1 if all constraints are satisfied and 0 otherwise.

As it is computationally expensive to enumerate all possible phrases and alignments during decoding, we resort to an external bilingual phrase table \cite{Koehn:03} to restrict the search space. Before decoding, the candidate translations of each source phrase, which are usually referred to as {\em translation options}, can be collected by matching the phrase table against the input sentence. Note that unlike \citet{Koehn:03}, our approach allows a source phrase or a target phrase to be unaligned.

It is easy for our approach to impose lexical constraints during the option collection process simply by replacing the translation of the pre-specified source phrase with the pre-specified target phrase. To achieve this, we restrict that (1) the pre-specified source phrase must be translated into a continuous segment and (2) its translation options do not overlap with other words. To impose structural constraints, we restrict that the translation options within a paired HTML tags do not intersect with those outside.

As unconstrained decoding is a special case of structurally constrained decoding and lexically constrained decoding can be achieved by restricting translation options, we focus on describing the structurally constrained decoding algorithm. We use a deductive system to formally describe the decoding process. An item in the deductive system is a 4-tuple $[\mathbf{x}, \mathbf{c}, S, \mathbf{y}]$ defined as follows: \footnote{As it is easy to obtain phrase alignment during the decoding process, we omit it in the item for simplicity.}
\begin{enumerate}

\item {\em Source sentence} $\mathbf{x}$: To capture structural constraints, we add {\em open constraint tags} (e.g., ``$\langle$c1$\rangle$'' and ``$\langle$c2$\rangle$'') and {\em close constraint tags} (e.g., ``$\langle$/c1$\rangle$'' and ``$\langle$/c2$\rangle$'') to the input, as shown in Figure \ref{fig:derivation}. Note that sentence boundaries can also be seen as constraints.

\item {\em Coverage vector} $\mathbf{c}$: A vector that consists of 0's and 1's to indicate which source words have been covered. The coverage vector is initialized as $\{0\}^I$.

\item {\em Stack} $S$: A stack that stores constraint tags. The decoding algorithm uses the stack to preserve structural constraints.

\item {\em Translation} $\mathbf{y}$: Partial translation generated during the decoding process.

\end{enumerate}

Each item is associated with a log probability $p$ yielded by our model. Note that a translation option can also be represented as an item $[\mathbf{x}, \mathbf{c}, \emptyset, \mathbf{y}]$. Except for the position of the source phrase, all other positions in $\mathbf{c}$ are set to 0. $\mathbf{y}$ is simply the target phrase. The log probability of a translation option is set to 0.

As shown in Figure \ref{fig:deductive}, the deductive system comprises three inference rules:

\begin{enumerate}

\item {\em Translate}: Translate a source phrase using a translation option. In Figure \ref{fig:deductive}, $[\mathbf{x}, \mathbf{c}_1, S, \mathbf{y}]$ is a current item and $[\mathbf{x}, \mathbf{c}_2, \emptyset, \mathbf{y}_2]$ is a translation option. This rule is activated in two cases: the translation option covers an uncovered source phrase within the constraint \footnote{By ``within the constraint'', we mean that the constraint is the innerest one that encloses a token.  For example, in Figure \ref{fig:derivation}, ``Edgar'' is within the inner constraint c2 rather than the outer constraint c1.} at the top of the stack, or the source phrase is empty (i.e., $\mathbf{c}_2 = \{0\}^I$). 

\item {\em Push}: Push a constraint tag to the stack. The algorithm constructs a special translation option $[\mathbf{x}, \mathbf{c}_2, \emptyset, s]$ for a constraint tag $s$. For the open constraint tag ``$\langle$c$\rangle$'', this rule is activated when all source words within the constraint are uncovered and the algorithm starts to translate any source phrase within the constraint. 
For the close constraint tag ``$\langle$/c$\rangle$'', this rule is activated when all source words within the constraint are covered.

\item {\em Pop}: Pop the top two constraint tags from the stack. This rule is activated if the top two elements in the stack are paired open and close tags (e.g., ``$\langle$c1$\rangle$'' and ``$\langle$/c1$\rangle$'').

\end{enumerate}

Similar to lexically constrained decoding \cite{Hokamp:17,Post:18}, we use an $I \times J$ matrix $\mathbf{M}$ to store all items generated during decoding, where $I$ is the length of input and $J$ is the maximum length of the output. Each element $M_{i, j}$ is a stack of items with $i$ source words covered and $j$ target words generated. While the time complexity of the decoding algorithm in standard NMT is $\mathcal{O}(bJ)$, the time complexity of our algorithm is $\mathcal{O}(bIJ)$, where $b$ is the beam size (i.e., the maximum number of items stored in each stack). To speed up the decoding, our approach only keeps top-$b$ items for all stacks with the same number of generated target words (i.e., $M_{*, j}$). As a result, the time complexity of our algorithm is reduced to $\mathcal{O}(kJ)$, which is identical to that of \citet{Post:18}.

\section{Experiments}

\subsection{Setup}
We evaluated our approach on the Chinese-English translation task. The training set contains 1.25M sentence pairs from LDC \footnote{The training set is composed of LDC2002E18, LDC2003E07, LDC2003E14, part of LDC2004T07, LDC2004T08, and LDC-2005T06.} with 29.8M Chinese tokens and 35.8M English tokens after byte pair encoding \cite{Sennrich:16} with 32K merges. The NIST 2006 dataset is used as the development set and the NIST 2008 datasets is used as the test set. The evaluation metric is case-insensitive BLEU4 \cite{Papineni:02} as calculated by the {\em multi-bleu.perl} script.

The NMT model used in our experiments is Transformer \cite{Vaswani:17}. The number of layers is set to 6 for both encoder and decoder. The hidden size is set to 512 and the filter size is set to 2,048. There are 8 separate heads in the multi-head attention. We used Adam \cite{Kingma:15} to optimize model parameters. During training, each batch contains approximately 25,000 tokens. We adopt the learning rate decay policy as described by \citet{Vaswani:17}. The length penalty \cite{Wu:16} is used and the hyper-parameter $\alpha$ is set to 0.6.

For our approach, we used the training set to train the non-empty translation model in Eq. (\ref{eq:non-empty}). The same training set was also used to obtain an aligned parallel corpus using GIZA++ \cite{Och:03}, which is used to extract a bilingual phrase table \cite{Koehn:03} to collect translation options and train the empty translation model in Eq. (\ref{eq:empty}).  The translation options of the empty source phrase are restricted to most frequent words of which the probabilities of aligning to the empty source phrase are higher than 0.2 on the training set.

\begin{table}
\centering
\begin{tabular}{|c|c||c|}
\hline
\multicolumn{2}{|c||}{empty} & \multirow{2}{*}{BLEU} \\
\cline{1-2}
source & target &  \\
\hline \hline 
$\times$ & $\times$ & 41.69 \\
$\times$ & $\surd$ & 47.43 \\
$\surd$ & $\times$ & 39.28 \\
$\surd$ & $\surd$ & \textbf{47.77} \\
\hline
\end{tabular}
\caption{Effect of empty phrases on the source and target sides on the development set.} \label{table:empty}
\end{table}

\begin{table}[!t]
\centering
\begin{tabular}{|c||c|c|}
\hline
model & MT06 & MT08 \\
\hline \hline
Transformer & 47.44 & 37.49 \\
\hline
{\em this work} & \textbf{47.77} & \textbf{38.14} \\
\hline
\end{tabular}
\caption{Comparison between the standard Transformer model and our latent variable model.} \label{table:final_results}
\end{table}

\subsection{Results on Unconstrained Decoding}

In this experiment, we compared our method with the standard Transformer model \cite{Vaswani:17}.

\subsubsection*{Effect of Empty Phrases}

Table \ref{table:empty} shows the effect of empty source and target phrases on the development set. The empty source phrase allows for target word insertion and the empty target phrase permits source word omission. It is clear that introducing empty phrases on both sides is beneficial for improving translation quality, suggesting that it is important to use empty phrases to reduce the discrepancy between the phrase-based search space and neural models. An interesting finding is that allowing for target word insertion but disabling source word omission dramatically hurts the translation performance (i.e., 39.28). We find that the decoder tends to insert many meaningless target words.

\subsubsection*{Comparison with Transformer}

Table \ref{table:final_results} shows the comparison between the standard Transformer model and our latent variable model. Our model is different from the standard model in two aspects. First, our model uses a phrase lattice to represent the search space. Second, empty phrases are introduced to make the search space more flexible than that of conventional SMT. We find that our model slightly improves over the standard model, suggesting that we can use the phrase-based search space to replace the standard search space for lexically and structurally constrained decoding. 

\begin{figure}[!t]
\centering
\includegraphics[width=0.48\textwidth]{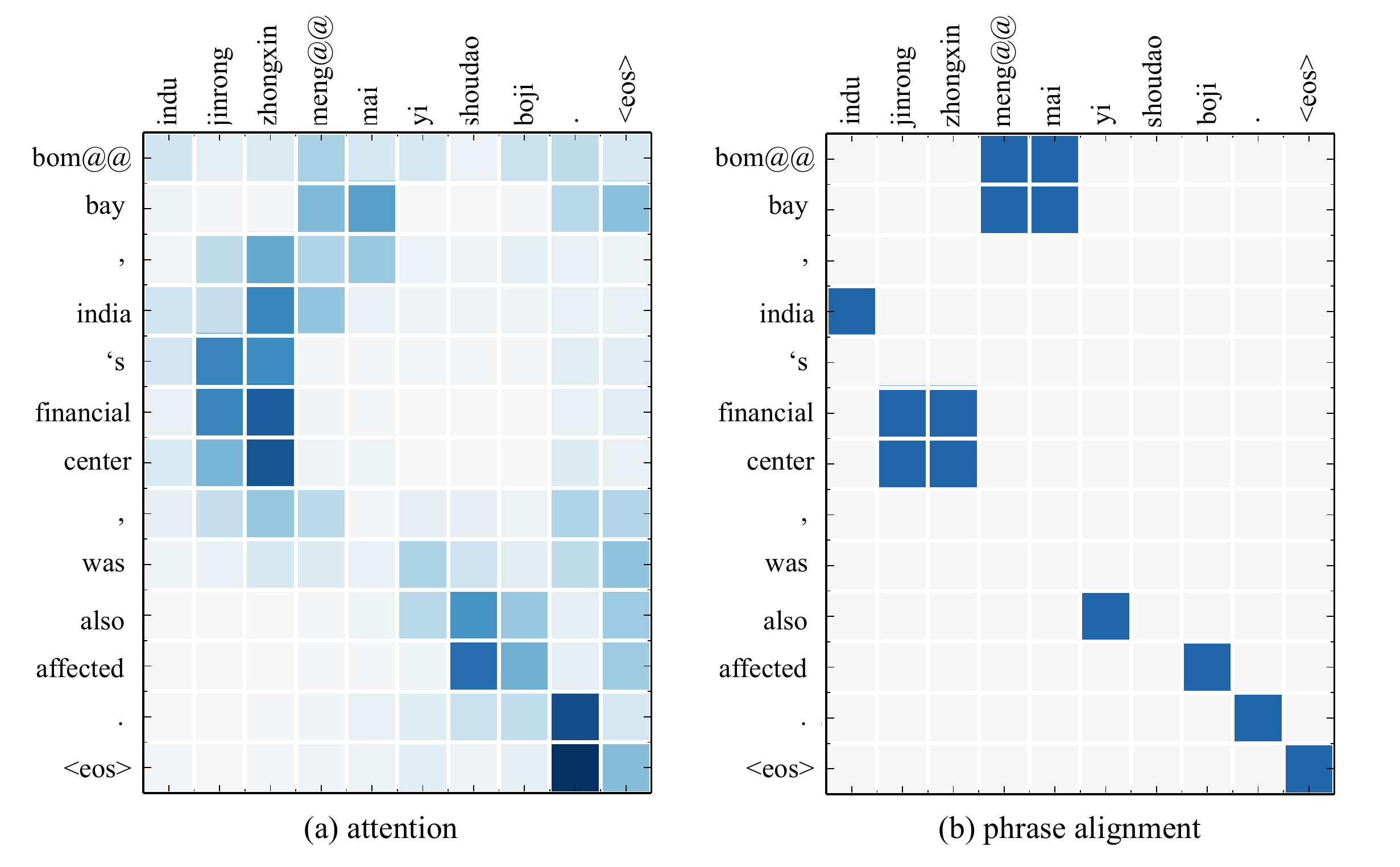}
\caption{Comparison between attention and alignment.} \label{fig:alignment}
\end{figure}

\begin{table*}[!t]
    \centering
    \begin{tabular}{|l|l|}
    \hline
    lexical constraints & (``\textcolor{blue}{taose}'', ``\textcolor{blue}{color blossoms}'') \\ [1pt]
    \hline
    source & `` { \textcolor{blue}{taose}} '' qian ban duan \textit{hai ting hao de} , dajia dou shi xinren . \\ [1pt]
    \hline 
    reference & the first half of ``{ \textcolor{blue}{color blossoms}} '' is quite good . they are all first-timers . \\ [1pt]
    \hline
    no constraint & the first half of the `` peach color '' is still quite good . people are new people . \\ [1pt]
    DBA & in the first half of the `` \textbf{peach} {\textcolor{blue}{color blossoms}} , '' people are new people . \\ [1pt]
    \textit{this work} & the first half of the `` {\textcolor{blue}{color blossoms}} '' is still quite good . people are new people . \\ [2pt]
    \hline \hline
    lexical constraints & (``\textcolor{blue}{yaoguanju}'', ``\textcolor{blue}{fda}''), (``\textcolor{red}{7 yue 30 ri}'', ``\textcolor{red}{july 30}''), (``\textcolor{orange}{wendiya}'', ``\textcolor{orange}{avandia}'') \\ [1pt]
    \hline
    source & \textcolor{blue}{yaoguanju} jiang yu  \textcolor{red}{7 yue 30 ri} juxing youguan \textcolor{orange}{wendiya} anquanxing de tingzhenghui \\ [1pt]
    \hline
    reference & the \textcolor{blue}{fda} will hold a hearing into the safety of \textcolor{orange}{avandia} on \textcolor{red}{july 30} . \\ [1pt]
    \hline
    no constraint & the drug administration will hold hearings on the safety of wendiya on \textcolor{red}{july 30} . \\ [1pt]
    DBA  & \textcolor{blue}{fda} \textcolor{orange}{avandia} will hold a hearing on the safety of \textbf{man dim} on \textcolor{red}{july 30}. \\ [1pt]
    \textit{this work} & the \textcolor{blue}{fda} will hold hearings on the safety of \textcolor{orange}{avandia} on \textcolor{red}{july 30} . \\ [1pt]
    \hline
    \end{tabular}
    \caption{Example translations of two lexically constrained decoding algorithms. We use DBA to denote the dynamic beam allocation method proposed by \cite{Post:18}. Lexical constraints are highlighted in different colors. We find that although DBA is able to include all specified target phrases in the translations, it tends to either translate the specified source phrases repeatedly (highlighted in bold) or omitting source phrases (highlighted in italic).}
    \label{tab:lex_con}
\end{table*}

\subsubsection*{Visualization}

Figure \ref{fig:alignment} shows the comparison between the attention and alignment. As there is no attention between the input and output in the Transformer model, the heatmap in Figure \ref{fig:alignment} is taken from the encoder-decoder attention in the third layer. In the heatmap, the attention weight is averaged over 8 different heads. While the attention matrix only reveals the relevance between source and target words, the phrase alignment generated by our model is more useful for achieving lexically and structurally constrained decoding.

\subsection{Results on Lexically Constrained Decoding}

In this experiment, we compared our method with dynamic beam allocation (DBA) proposed by \citet{Post:18}. We asked human experts to pre-specify 467 distinct lexical constraints with 1,005 occurrences for the NIST 2008 dataset. They are mostly translations of named entities.

We find that imposing lexical constraints using DBA achieves a BLEU score of 38.54 and our approach achieves a BLEU score of 39.43. Table \ref{tab:lex_con} shows some example translations. Given a lexical constraint (``taose'', ``color blossoms''), unconstrained decoding fails to generate ``color blossoms'' on the target side. DBA is capable of enforcing the target phrase of the lexical constraint to appear in the translation. However, there is an extra target word ``peach'' (highlighted in bold) that is also connected to ``taose''. In other words, ``taose'' is translated twice in a wrong way. To make things worse, DBA omits the source phrase ``hai ting hao de'' (highlighted in italic). Similar findings are also observed on the second example, in which the Chinese word ``wendiya'' is translated twice by DBA: ``avandia'' and ``man dim'' (highlighted in bold). 

We observe that 6.9\% of the source phrases of lexical constraints on the test set are repeatedly translated by DBA while the proportion drops to 0.3\% for our approach. One possible reason is that DBA ignores the source side of a lexical constraint and thus inevitably impairs the adequacy of the resulting translation.

\subsection{Results on Structurally Constrained Decoding}

We evaluated our structurally constrained decoding algorithm on a webpage translation task.

\subsubsection*{Dataset}

As labeled data is limited in quantity for webpage translation, we still use the unstructured Chinese-English dataset that contains 1.25M sentence pairs as the training set. We built a test set for Chinese-English structured text translation based on the webpages of Wikipedia. The test set contains 200 sentences with HTML tags retained. On average, each sentence in the test set has 36.9 words and 2.6 pairs of HTML tags.

\subsubsection*{Baselines}
We compared our approach with the following five baselines: \footnote{We did not compare with the methods that train SMT models on parallel corpora for webpage translation because these datasets are not publicly available.}

\begin{enumerate}
\item \textproc{Remove}: Remove all HTML tags before decoding and do not insert tags back to translations after decoding. 
\item \textproc{Split} \cite{Al-Anzi:97}: Split the input by tags before decoding, translate textual parts independently, and concatenate translations monotonically after decoding.
\item \textproc{Match} \cite{Zantout:01}: Remove all HTML tags before decoding and insert tags back to translations by matching.
\item \textproc{Align} \cite{Joanis:13}: Remove all HTML tags before decoding and insert tags back to translations using word alignments generated by GIZA++.
\item \textproc{Google}: The Google Translate online system. \footnote{\url{https://translate.google.com/}}
\end{enumerate}

All the baselines except \textproc{Google} share the same Transformer model with our approach. 

\begin{table}[!t]
\centering
\begin{tabular}{|l|c|c|c|}
\hline 
Method & w/o tag & w/ tag & in tag \\
\hline \hline
\textproc{Remove} & 33.29 & 22.08 & - \\
\textproc{Split}  & 20.39 & 24.54 & 37.66 \\
\textproc{Match} & 33.29 & 33.32 & 37.74 \\
\textproc{Align} & 33.29 & 36.71 & 30.98 \\
\hline
\textproc{Google} & 31.30 & 35.36 & 34.41 \\
\hline \hline
Ours & \textbf{33.42} & \textbf{37.52} & \textbf{38.41} \\ 
\hline
\end{tabular}
\caption{Results on the webpage translation task. ``w/o tag'' denotes the BLEU score without considering HTML tags, ``w/ tag'' denotes the BLEU score considering HTML tags, and ``in tag'' denotes the BLEU score considering only the text enclosed by tags. Note that \textproc{Remove} does not have the ``in tag'' BLEU score because all tags are removed.} \label{table:structred_result}
\end{table}

\subsubsection*{Results on Webpage Translation}
Table \ref{table:structred_result} shows the comparison of imposing structural constraints with existing methods on the test set. As \textproc{Remove} ignores all HTML tags, it is not capable of imposing structural constraints. \textproc{Split} ensures that the structural constraints can be imposed correctly because the sentence segments between HTML tags are translated independently, but the translation quality drops dramatically. \textproc{Match} and \textproc{Align} take the full advantage of standard NMT to translate the textual parts but often fail to recover HTML tags correctly after decoding. According to the translations, it seems that \textproc{Google} uses a strategy similar to \textproc{Split} but achieves much higher BLEU scores because it used much larger training data than all other methods. Our approach achieves the best performance in terms of all evaluation metrics by fully preserving the structural constraints without losing translation quality.

\section{Conclusion}

We have presented a latent variable model for neural machine translation that treats phrase alignment as an unobserved latent variable. The introduction of phrase alignment makes it possible to decompose the translation process of arbitrary NMT models into interpretable steps. In addition, it is also convenient to use our approach to impose lexical and structural constraints thanks to the availability of phrase alignment. Experiments show that the proposed method achieves significant better performance on both lexically and structurally constrained translation tasks. 

\bibliography{aaai2020_zjc}
\bibliographystyle{acl_natbib}

\end{document}